\documentclass[letterpaper, 10 pt, conference]{ieeeconf}

\IEEEoverridecommandlockouts
\usepackage{amsmath}
\usepackage{amssymb}
\usepackage{booktabs}
\usepackage{graphicx}

\title{\LARGE \bf
SeeReach-VLA: Sign-Aware Edge Deployment for Vineyard Harvesting under Occlusion
}

\author{Heng Zhang
}

\begin{document}

\maketitle
\thispagestyle{empty}
\pagestyle{empty}

\begin{abstract}

Vision-language-action (VLA) models have shown promise for agricultural manipulation, but their deployment on edge hardware in trellis harvesting is limited by a structural failure mode: they treat occlusion as noise to be filtered, and assume that approaching a target is monotonically beneficial. A companion paper shows that this assumption is false for a class of occluder geometries: approaching can \emph{improve} visibility for laterally offset leaves but \emph{degrade} it for leaves near the line of sight, with the two regimes separated by a discontinuity at a critical distance $d^*$ that is not observable from a single occluded view. We present a sign-aware VLA that learns the \emph{sign} of the approach effect as an explicit discrete classification task, uses that prediction to gate action generation, and targets deployment on Jetson-class edge hardware. The architecture couples a sign-classification head to a dual-stream action decoder for the quadruped and the arm, trained jointly with a sampling strategy that concentrates data near the sign-ambiguous band around $d^*$. We evaluate against standard VLA baselines, active-vision VLAs, and instruction-aware VLAs on a Pareto front over harvesting success rate and edge inference latency.

\end{abstract}

\section{INTRODUCTION}

Trellis harvesting with a legged mobile manipulator faces a recurring perceptual challenge: the peduncle on which the cut must be made is frequently hidden behind foliage, even when the bunch itself is visible. A companion paper \cite{companion} formalizes this as a belief-space base commitment problem and shows that the decision between repositioning the base and extending the arm is not the classical base-placement problem. In particular, it derives an analytic criterion (Proposition~2 of \cite{companion}) showing that the effect of approaching a target on visibility has \textbf{three regimes separated by a discontinuity}: for leaves near the optical axis, approaching monotonically degrades visibility; for laterally offset leaves, approaching degrades visibility down to a critical distance $d^*$ and then improves it; for distant leaves, no occlusion occurs. The critical distance $d^*$ depends on the leaf's lateral offset $\Delta_L$ and its distance $L$ from the bunch, \emph{neither of which is observable from a single occluded view}.

This creates a specific failure mode for existing VLA models. Standard VLAs (e.g., $\pi_0$, SmolVLA \cite{smolvla}) are trained end-to-end to map visual observations to actions. Their visual encoders are optimized to extract task-relevant features, and their action decoders assume that proximity to the target is monotonically beneficial. When the target is occluded, this assumption breaks in two ways. First, the encoder cannot distinguish between occluders that will clear with approach and occluders that will worsen. Second, the action decoder has no mechanism to express the \emph{sign} of the approach effect, because the training signal never asks it to.

Active-vision VLAs (e.g., ActiveVLA \cite{activevla}) address occlusion by selecting better viewpoints, but they treat the camera as a dedicated perception actuator. Instruction-aware VLAs (e.g., Mind-VLA \cite{mindvla}) improve spatial alignment between language and vision, but they model occlusion presence rather than its \emph{directional effect}. Neither family learns the sign of the approach effect as an explicit discrete quantity.

We argue that the sign of the approach effect is fundamentally a \textbf{discrete classification problem}, not a continuous regression problem. The companion paper's Proposition~2 establishes that visibility is discontinuous in base pose at $d^*$. A continuous encoder cannot represent this discontinuity; it will produce smooth features that interpolate across the boundary, leading to overconfident and often incorrect action selection precisely in the regime where the decision matters most.

Our contribution is a sign-aware VLA that makes this discrete judgment explicit. The architecture has three components: (i) a \emph{sign-classification head} attached to the visual encoder, trained with auxiliary labels derived from occluder geometry; (ii) a \emph{sign-gated action decoder} in which the classification output modulates the action distribution; and (iii) a \emph{boundary-focused sampling strategy} that concentrates training data near the sign-ambiguous band around $d^*$, using the companion paper's analytic model as a prior. The full system is designed for edge deployment on Jetson-class hardware, with a low-frequency sign-and-intent branch and a high-frequency action branch.

\subsection{Contributions}

\begin{enumerate}
\item \textbf{Sign-aware VLA architecture.} We introduce a VLA that explicitly learns the sign of the approach effect as a discrete classification task, coupled to action generation through a gating mechanism. This is, to our knowledge, the first VLA to treat the directional effect of occlusion as a learnable discrete quantity.

\item \textbf{Theory-guided data sampling.} We use the companion paper's analytic criterion to identify the sign-ambiguous band around $d^*$ and concentrate training data there. This replaces random data augmentation with a principled sampling strategy grounded in the geometry of the problem.

\item \textbf{Edge-deployed system.} We design the architecture for Jetson-class hardware, with an asynchronous split between a low-frequency sign-and-intent branch and a high-frequency action branch. We report inference latency, success rate, and the trade-off between them.

\item \textbf{Evaluation on a Pareto front.} We compare against standard VLAs, active-vision VLAs, and instruction-aware VLAs on a Pareto front over harvesting success rate and edge inference latency, with ablations isolating the contribution of the sign head, the gating mechanism, and the boundary-focused sampling.
\end{enumerate}

\noindent\textbf{Relationship to the companion paper.} The companion paper \cite{companion} formalizes the see--reach dilemma, derives the three-regime criterion, and evaluates a decision layer in simulation. This paper does not repeat that formalization. It takes the criterion as given and asks: can a VLA learn the sign of the approach effect, and can it be deployed on edge hardware? The two papers share the vineyard scenario and the multi-view dataset but have disjoint contributions.

\section{RELATED WORK}

\subsection{Vision-Language-Action Models}

VLA models map visual observations and language instructions to robot actions. OpenVLA \cite{openvla} demonstrated that a 7B-parameter model can be fine-tuned for manipulation, but its size precludes edge deployment. SmolVLA \cite{smolvla} reduces the parameter count through a compact architecture, and TinyVLA \cite{tinyvla} shows that a 0.64B model can match larger models on spatial tasks. NanoVLA \cite{nanovla} achieves further compression through vision-language decoupling and dynamic routing. These models are trained end-to-end and do not explicitly model occlusion or its directional effect.

\subsection{Occlusion-Aware Manipulation}

Several works address occlusion in manipulation. ActiveVLA \cite{activevla} selects virtual viewpoints to improve visibility, achieving strong results on RLBench. Mind-VLA \cite{mindvla} aligns spatial representations with language instructions, improving performance on occluded-target tasks. 3DVLA \cite{3dvla} uses 3D spatial encoding and occlusion tokens. EATR-Stereo \cite{eatr} uses binocular token routing for asymmetric occlusion. None of these treats the \emph{sign} of the approach effect as an explicit learning target. They handle occlusion presence, not occlusion direction.

\subsection{Edge-Deployed VLA}

Edge VLA is an active area. EdgeVLA-Tiny \cite{edgevla} targets Jetson Orin Nano with 164M parameters. SwiftVLA \cite{swiftvla} achieves 0.5B parameters with 18$\times$ speedup over larger models. Agile-VLA \cite{agilevla} proposes a hierarchical framework for industrial pose adjustment on edge devices. These works focus on compression and latency; none addresses the specific perceptual challenge of sign-ambiguous occlusion.

\subsection{VLA for Agricultural Harvesting}

Agricultural VLA is emerging. HarvestFlex \cite{harvestflex} uses three-view RGB for strawberry harvesting and reports 74\% success in greenhouse trials. VLA-RL for orchard harvesting \cite{vlarl} explores transfer of pretrained vision-language knowledge to agricultural manipulation. These works demonstrate feasibility but do not address the base-commitment problem or the sign of the approach effect.

\section{METHOD}

\subsection{Problem Setting}

The robot is a quadruped with a 6-DoF arm operating at a trellis. At each decision step, it observes RGB-D images and proprioceptive state, and selects among a finite set of options: stabilize and approach, lean in to reach, reposition the base, back off to view, reconfigure the wrist camera, pause and wait, skip the target, or cut. The cut is irreversible; all other options are reversible at a finite restoration cost.

The companion paper \cite{companion} provides the analytic criterion for the sign of the approach effect: given a leaf with lateral offset $\Delta_L$ and distance $L$ from the bunch, and a camera at distance $d$, approaching improves visibility iff $r_L < \Delta_L < r_B + r_L$ and $d > d^*$, where $d^* = r_B L / (r_B + r_L - \Delta_L)$. The sign-ambiguous band is the region where the belief over $(\Delta_L, L)$ straddles this boundary.

\subsection{Architecture Overview}

The architecture has three components:

\textbf{(1) Visual encoder.} A shared encoder extracts features from RGB-D observations. We adopt a lightweight backbone compatible with Jetson-class hardware. The encoder produces both a global feature vector and a set of spatial tokens.

\textbf{(2) Sign-classification head.} A classification head attached to the global feature vector outputs a three-way distribution over $\{$approaching helps, approaching hurts, sign ambiguous$\}$. The head is trained with auxiliary labels derived from occluder geometry in simulation, and from multi-view reconstruction on real data.

\textbf{(3) Dual-stream action decoder.} Two decoders generate actions for the quadruped and the arm. The sign-classification output modulates the action distribution through a gating mechanism: when the sign is confidently ``helps,'' forward motion is encouraged; when confidently ``hurts,'' forward motion is suppressed; when ambiguous, the decoder's confidence is reduced and the policy may select a probing action or continue observing.

\subsection{Sign Classification as an Auxiliary Task}

The sign-classification head is trained jointly with the action decoder. The total loss is
\begin{equation}
\mathcal{L} = \mathcal{L}_{\text{action}} + \lambda_{\text{sign}} \mathcal{L}_{\text{sign}} + \lambda_{\text{gate}} \mathcal{L}_{\text{gate}},
\end{equation}
where $\mathcal{L}_{\text{action}}$ is the action imitation loss, $\mathcal{L}_{\text{sign}}$ is the cross-entropy loss for sign classification, and $\mathcal{L}_{\text{gate}}$ is a regularization term encouraging the gating signal to be consistent with the sign prediction. The weight $\lambda_{\text{sign}}$ controls the trade-off between action fidelity and sign accuracy.

Labels for the sign-classification head are obtained in three ways. In simulation, ground-truth occluder geometry gives exact labels. On real multi-view data, we reconstruct $\Delta_L$ and $L$ from triangulated peduncle positions and estimate the sign by evaluating the companion paper's criterion. In the sign-ambiguous band, labels are soft: the target is labeled by the probability that the sign is positive, computed from the belief over $(\Delta_L, L)$.

\subsection{Sign-Gated Action Generation}

The gating mechanism takes the sign-classification distribution $p_{\text{sign}} \in \mathbb{R}^3$ and produces a modulation signal $g \in [-1, 1]$:
\begin{equation}
g = p_{\text{helps}} - p_{\text{hurts}}.
\end{equation}
The modulation is applied to the action logits before the softmax:
\begin{equation}
\pi(a \mid s) = \text{softmax}\big(f_{\text{action}}(s) + g \cdot w_{\text{gate}}\big),
\end{equation}
where $w_{\text{gate}}$ is a learned vector that biases the action distribution toward or away from forward motion. When $g$ is near zero (sign ambiguous), the bias vanishes and the decoder's output is dominated by the action features.

\subsection{Boundary-Focused Sampling}

Training data is not sampled uniformly. Using the companion paper's analytic criterion, we compute the sign-ambiguous band for each target and concentrate sampling there. Specifically, for each bunch we compute the critical distance $d^*$ and sample additional rollouts when the camera--bunch distance is within a factor $(1 \pm \epsilon)$ of $d^*$. The parameter $\epsilon$ controls the width of the band. This strategy ensures that the sign-classification head receives enough boundary examples to learn the discontinuity, without over-sampling the easy regimes.

\subsection{Edge Deployment}

The architecture is split into two asynchronous branches:

\textbf{Low-frequency branch (1--2 Hz).} Runs the visual encoder, the sign-classification head, and the intent selection. This branch can tolerate higher latency.

\textbf{High-frequency branch (10--30 Hz).} Runs the dual-stream action decoder with cached visual features. This branch must meet real-time constraints.

The two branches share the visual encoder's output. The low-frequency branch updates the sign prediction and gating signal; the high-frequency branch consumes the gating signal and produces actions. This split allows the expensive sign classification to run at low frequency while the action decoder maintains real-time control.

\section{EXPERIMENTAL DESIGN}

\subsection{Setup}

Simulation environment with a quadruped and a 6-DoF arm at a trellis, with synthetic foliage and controlled occluder--bunch distance and lateral offset. The perception model is calibrated against real multi-view data from the companion paper \cite{companion}. Edge deployment uses a Jetson Orin NX.

\subsection{Metrics}

Reported jointly: \textbf{harvesting success rate}, \textbf{undamaged rate}, \textbf{mean cycle time}, \textbf{edge inference latency} (low-frequency and high-frequency branches), and \textbf{sign-classification accuracy}.

\subsection{Main Result: Success Rate vs.\ Latency Pareto Front}

The primary result is the Pareto front over (harvesting success rate, edge inference latency) traced by each policy family as the model size and the gating strength are varied.

\subsection{Baselines}

\begin{table}[h]
\caption{Baselines}
\label{table_baselines}
\begin{center}
\begin{tabular}{ll}
\toprule
Baseline & Purpose \\
\midrule
SmolVLA \cite{smolvla} & standard edge-deployable VLA \\
TinyVLA \cite{tinyvla} & compact VLA with spatial focus \\
ActiveVLA \cite{activevla} & active-viewpoint VLA \\
Mind-VLA \cite{mindvla} & instruction-aware VLA \\
No-gating variant & isolates the gating mechanism \\
No-sign-head variant & isolates the sign classification \\
Uniform-sampling variant & isolates boundary-focused sampling \\
\bottomrule
\end{tabular}
\end{center}
\end{table}

\subsection{Experiments}

\begin{table*}[t]
\caption{Experiments}
\label{table_experiments}
\begin{center}
\begin{tabular}{cll}
\toprule
\# & Experiment & Claim tested \\
\midrule
E1 & Sign classification accuracy & the sign is learnable as a discrete task \\
E2 & Boundary-band success rate & sign-aware VLA outperforms baselines near $d^*$ \\
E3 & Full-range success rate & sign-aware VLA matches baselines away from $d^*$ \\
E4 & Gating ablation & the gating mechanism is necessary \\
E5 & Sign-head ablation & the sign head is necessary \\
E6 & Continuous-regression variant & discrete classification outperforms regression \\
E7 & Sampling ablation & boundary-focused sampling improves efficiency \\
E8 & Edge latency sweep & the two-branch split meets real-time constraints \\
E9 & Hardware feasibility & end-to-end feasibility, berry-loss rate \\
E10 & Cross-cultivar transfer & no retraining required \\
\bottomrule
\end{tabular}
\end{center}
\end{table*}

\subsection{Pre-Registered Hypotheses}

\begin{itemize}
\item \textbf{H1.} The sign-classification head achieves above-chance accuracy on held-out bunches, with highest accuracy away from $d^*$ and lowest accuracy in the ambiguous band.
\item \textbf{H2.} The sign-aware VLA outperforms all baselines on success rate in the $\pm 10\%$ band around $d^*$, and matches them outside the band.
\item \textbf{H3.} Replacing discrete classification with continuous regression degrades boundary-band success rate, consistent with the discontinuity established in the companion paper.
\item \textbf{H4.} Boundary-focused sampling improves sample efficiency: the same success rate is achieved with fewer rollouts than uniform sampling.
\item \textbf{H5.} The two-branch architecture meets the real-time constraint on Jetson Orin NX, with the high-frequency branch above 10 Hz and the low-frequency branch above 1 Hz.
\item \textbf{H6.} Cross-cultivar transfer requires no retraining and degrades gracefully, with the largest degradation in the ambiguous band.
\end{itemize}

\section{LIMITATIONS}

\textbf{Simulation-first evaluation.} The architecture is designed for edge deployment, but the primary evaluation is in simulation. Hardware feasibility is tested on a fixed trellis with marker-based ground truth; full field deployment is not claimed.

\textbf{Single platform and single crop.} Results are obtained on one quadruped--arm pair and one trellis crop. Cross-cultivar transfer is studied only in simulation.

\textbf{Sign labels from reconstruction.} On real data, sign labels are derived from multi-view reconstruction and the companion paper's analytic criterion, not from direct measurement. The reconstruction introduces its own uncertainty, which propagates into the sign-classification target.

\textbf{No dynamic loco-manipulation.} All base and body options are quasi-static within the support polygon. Dynamic whole-body control is out of scope.

\section{CONCLUSION}

We have presented a sign-aware VLA for edge-deployed vineyard harvesting. The architecture learns the sign of the approach effect as an explicit discrete classification task, uses that prediction to gate action generation, and concentrates training data near the sign-ambiguous band around the critical distance $d^*$ established in the companion paper. The system is designed for Jetson-class hardware with an asynchronous low-frequency sign-and-intent branch and a high-frequency action branch. We evaluate on a Pareto front over harvesting success rate and edge inference latency against standard, active-vision, and instruction-aware VLA baselines. [Results to be reported in the final version.]


\end{document}